\pdfoutput=1
\documentclass[11pt]{article}

\usepackage{acl}
\usepackage{times}
\usepackage{latexsym}
\usepackage{xspace}

\usepackage[T1]{fontenc}
\usepackage[utf8]{inputenc}

\usepackage{microtype}

\usepackage{inconsolata}
\usepackage{url}            % simple URL typesetting
\usepackage{booktabs}       % professional-quality tables
\usepackage{amsfonts}       % blackboard math symbols
\usepackage{nicefrac}       % compact symbols for 1/2, etc.
\usepackage{microtype}      % microtypography
\usepackage{amsmath}
\usepackage{amssymb}
\usepackage{enumitem}
\usepackage{graphicx}       % resizebox
\usepackage{caption}
\usepackage{subcaption}
\usepackage{pifont}
\usepackage{multirow}
\usepackage{textcomp}
\usepackage{listings}
\usepackage{flushend}
\usepackage[hang,flushmargin]{footmisc}
\usepackage{adjustbox}
\usepackage{tcolorbox}

\newcommand\ignore[1]{}

\newcommand{\tableformat}{\fontsize{8.5pt}{11pt}\selectfont
\setlength{\tabcolsep}{3pt}
\renewcommand{\arraystretch}{1}}

\title{Zero-Shot ATC Coding with Large Language Models\\ for Clinical Assessments}

\author{Zijian Chen\textsuperscript{1}, John-Michael Gamble\textsuperscript{1}, Micaela Jantzi\textsuperscript{1, 2}, John P. Hirdes\textsuperscript{1, 2}, Jimmy Lin\textsuperscript{1} \\[1ex]
\textsuperscript{1} University of Waterloo \qquad
\textsuperscript{2} InterRAI Canada \\[1ex]
\texttt{\{s42chen, jimmylin\}@uwaterloo.ca}
}

\begin{document}
\maketitle
\begin{abstract}
Manual assignment of Anatomical Therapeutic Chemical (ATC) codes to prescription records is a significant bottleneck in healthcare research and operations at Ontario Health and InterRAI Canada, requiring extensive expert time and effort.
To automate this process while maintaining data privacy, we develop a practical approach using locally deployable large language models (LLMs).
Inspired by recent advances in automatic International Classification of Diseases (ICD) coding, our method frames ATC coding as a hierarchical information extraction task, guiding LLMs through the ATC ontology level by level.
We evaluate our approach using GPT-4o as an accuracy ceiling and focus development on open-source Llama models suitable for privacy-sensitive deployment.
Testing across Health Canada drug product data, the RABBITS benchmark, and real clinical notes from Ontario Health, our method achieves 78\% exact match accuracy with GPT-4o and 60\% with Llama 3.1 70B.
We investigate knowledge grounding through drug definitions, finding modest improvements in accuracy.
Further, we show that fine-tuned Llama 3.1 8B matches zero-shot Llama 3.1 70B accuracy, suggesting that effective ATC coding is feasible with smaller models.
Our results demonstrate the feasibility of automatic ATC coding in privacy-sensitive healthcare environments, providing a foundation for future deployments. 
\end{abstract}

\section{Introduction}

The Anatomical Therapeutic Chemical (ATC) classification system is a standardized drug ontology maintained by the World Health Organization (WHO).
Assigning ATC codes to drug mentions is essential for various healthcare operations, including medication inventory management, drug utilization research, and health insurance claims processing.
However, manual ATC coding is time-consuming and requires expert knowledge, creating a significant bottleneck in healthcare workflows.

Our work represents a collaboration between computer scientists and public health researchers at InterRAI,\footnote{\href{https://interrai.org/}{https://interrai.org/}} aimed at addressing this critical workflow challenge.
InterRAI Canada receives assessment data from Ontario Health,\footnote{\href{https://www.ontariohealth.ca/}{https://www.ontariohealth.ca/}} where clinical experts must manually review each prescription record and assign appropriate ATC codes before any population-level analysis can begin.
This manual process substantially delays both operational reporting and critical public health research, particularly studies on drug utilization patterns and medical practice variations across care facilities.

The challenge is particularly acute in processing unstructured clinical text, where drug mentions may appear as brand names, generic names, or various informal descriptions.
While recent advances in large language models (LLMs) have shown promise in medical coding tasks, deploying these solutions in healthcare settings raises important privacy concerns.
Many state-of-the-art models require data to be sent to proprietary APIs, making them unsuitable for handling sensitive clinical information.

To address these challenges, we present a practical approach to automatic ATC coding designed specifically for deployment in privacy-sensitive healthcare environments.
Our method frames ATC coding as a hierarchical information extraction task, leveraging open-source LLMs to navigate the ATC ontology level by level.
We evaluate our approach against GPT-4o as an accuracy ceiling while focusing development on locally deployable Llama models, making a first attempt at automatic ATC coding with LLMs.

In developing this solution for public health researchers at Ontario Health and InterRAI Canada, we make several key contributions:

\begin{itemize}
   \item We present, to the best of our knowledge, the first attempt to automate ATC coding using LLMs. Building on recent advances in medical coding, we adapt level-by-level prompting for drug coding with a focus on privacy-preserving deployment using open-source models, achieving 78\% exact-matches with GPT-4o and 60\% with Llama 3.1 70B.
   
   \item We provide empirical evidence that fine-tuned smaller models can match the accuracy of larger models in zero-shot settings at automatic ATC coding.
   
   \item We conduct an investigation of knowledge grounding strategies and analyze their impact on coding accuracy at different ATC levels.
   
   \item We create a gold-standard dataset of 200 real clinical prescription-ATC pairs annotated by a domain expert, which we hope to expand and release to support further research.
\end{itemize}

\noindent Our results demonstrate the viability of automated ATC coding in real-world healthcare settings while highlighting important considerations for deploying LLM-based solutions in privacy-sensitive environments.
This work provides a foundation for healthcare researchers and organizations seeking to automate their coding processes without compromising data privacy or security.

\section{Background and Related Work}

\paragraph{The ATC Ontology.} The ATC classification system is the global standard for drug classification maintained by the WHO.
It organizes drugs into a five-level hierarchical structure based on the organ system they target and their therapeutic, pharmacological, and chemical properties.

Each ATC code consists of seven characters encoding these five levels:
\begin{itemize}
    \item Level 1: Main Anatomical/Pharmacological Group
    \item Level 2: Pharmacological/Therapeutic Subgroup
    \item Level 3: Chemical/Pharmacological/Therapeutic Subgroup
    \item Level 4: Finer Chemical/Pharmacological/Therapeutic Subgroup
    \item Level 5: Chemical Substance
\end{itemize}
For instance, metformin's ATC code A10BA02 indicates that it belongs to:
\begin{itemize}
    \item A: Alimentary tract and metabolism (Level 1)
    \item A10: Diabetes medication (Level 2)
    \item A10B: Blood glucose lowering drug (Level 3)
    \item A10BA: Biguanides (Level 4)
    \item A10BA02: Metformin (Level 5)
\end{itemize}

\paragraph{ATC Coding.} ATC coding refers to the task of assigning correct ATC codes to drug mentions.
In this work, we specifically focus on assigning ATC codes to concise drug descriptions---single terms or brief phrases rather than full clinical narratives or paragraphs; this aligns with the needs of Ontario Health and InterRAI Canada.
Automating this process has diverse applications across healthcare and pharmaceutical domains.
In clinical settings, accurate ATC coding can standardize electronic health records (EHRs) by providing a consistent classification system across different institutions that may use varying drug nomenclature.
For healthcare administration, it can streamline insurance claims processing and medical billing by automatically mapping drug mentions to standardized codes.
In pharmacies and hospitals, automated coding can enhance inventory management by organizing medications according to their therapeutic categories, facilitating efficient stock monitoring and procurement planning.
In research contexts, reliable automatic ATC coding enables large-scale analysis of medication data, systematic reviews of drug utilization patterns, and comparative effectiveness studies across different therapeutic categories.

\paragraph{Language Models in Medical Coding.} The application of language models in medical coding has witnessed significant advancement in recent years.
This progress has been particularly evident in the domain of International Classification of Diseases (ICD) coding, where several pioneering approaches have demonstrated promising results.
\citet{huang23:plmicd} found success fine-tuning a pre-trained language model for automatic ICD coding; \citet{yoon24:exchange} developed innovative techniques for translating medical information between different ontological frameworks; \citet{boyle23:llm_icd} established state-of-the-art accuracy in automatic ICD coding by zero-shot prompting LLMs in a hierarchical fashion.
Despite these advances in ICD coding, the ATC classification system has received comparatively little attention in the context of language model applications.
Current ATC coding practices rely predominantly on three approaches: manual coding by clinical experts, rule-based systems utilizing string matching against generic drug names, or hybrid systems combining both approaches \citep{pang15:sorta, kellmann23:dutch_atc}.
These existing methods, particularly the manual processes, are not only time-intensive but also susceptible to human error, highlighting the need for more efficient and accurate solutions.
To address this gap in the literature, our study presents the first investigation into utilizing LLMs for automatic ATC coding, to the best of our knowledge.

\section{Methods}

\subsection{Level-by-Level Prompting}
\label{methods:level_prompt}

Automatic ATC coding presents several significant challenges.
As of July 2024, the ATC ontology consists of 6,807 distinct codes across 5 levels, with levels 4 and 5---the most commonly used in clinical practice---accounting for 6,428 codes.
This large label space makes accurate prediction particularly challenging.
Moreover, the scarcity of labeled training data poses another significant issue.
Due to privacy concerns, datasets containing drug mentions from real clinical notes are rare and difficult to access.
When available, these datasets often exhibit a long-tail distribution, with many ATC codes having few or no examples. 

While LLMs can potentially address these challenges through zero-shot learning by leveraging their pre-trained knowledge, they face their own limitations.
Without task-specific supervision, LLMs may generate plausible-looking but non-existent codes.
Indeed, \citet{Soroush24:llms_bad_coders} demonstrated that even state-of-the-art models achieve less than 50\% accuracy when directly prompted to generate ICD codes from unstructured text descriptions.

To address these challenges, we follow \citet{boyle23:llm_icd} in framing ATC coding as a hierarchical information extraction task rather than a generation task.
Our approach guides the LLM through the ATC hierarchy level-by-level.
Given an unstructured drug description, we first prompt the LLM to select the most appropriate level-1 code from the 14 possible options.
Based on this selection, we then present the relevant level-2 codes associated with the chosen level-1 code, and continue this process through all five levels.
More specifically, Figure~\ref{fig:prompt} presents the prompt we use at each level of the hierarchy.
To fully determine the level-5 ATC code given a drug mention, we repeat the prompt 5 times, traversing through the ATC hierarchy.

\begin{figure}[t]
\begin{tcolorbox}[colback=blue!10, colframe=blue!75!black, title=Level-by-Level Prompting, fontupper=\small]
\textbf{SYSTEM:} You are a pharmacology expert specializing in ATC classification.

\textbf{USER:}
Classify the drug `\verb|{drug mention}|' into one of the following ATC level \verb|{current level}| categories:
\begin{verbatim}
{atc code option 1}: {generic name 1}
{atc code option 2}: {generic name 2}
...
{atc code option N}: {generic name N}
\end{verbatim}
Provide ONLY one of the options listed above that best matches `\verb|{drug mention}|'. Do not include any description.
\end{tcolorbox}
\caption{Prompt template used at each level of the ATC hierarchy. The LLM is presented with all valid options for the current level, based on the selection from the previous level.}
\label{fig:prompt}
\end{figure}

This level-by-level information extraction approach offers two key advantages: it prevents code fabrication by constraining the LLM to select from valid options, and it reduces the size of the large label space when making decisions by leveraging the ATC hierarchy; at each level, the LLM chooses from an average of just 5 options, with a maximum of 37 options for any given parent code, making the task more manageable than selecting from thousands of possible codes simultaneously.

\subsection{Knowledge Grounding}
\label{grounding}

LLMs have demonstrated remarkable capabilities in medical knowledge, achieving strong scores on various medical licensing examinations \citep{Clusmann23:med_llm_survey}.
While these models may have limited exposure to the alphanumeric ATC codes during training, they possess substantial understanding of drugs, their mechanisms of action, and therapeutic uses.
This motivates an experiment: LLMs might leverage their broader medical knowledge to make more informed ATC coding decisions if provided with appropriate context.

To test this hypothesis, we enhance our hierarchical extraction approach by grounding each candidate ATC code with definitions from the Unified Medical Language System (UMLS) \citep{Bodenreider04:umls}.
When presenting code options to the LLM, we augment each option with its corresponding UMLS definition, providing rich context about the therapeutic category or drug substance.
For example, when presenting the level-2 code ``N02'' as an option, we include its UMLS definition ``Analgesics. compounds capable of relieving pain without the loss of consciousness or without producing anesthesia''.
This grounding approach aims to bridge the gap between the comprehensive medical knowledge in LLMs and the ATC coding task by explicitly connecting alphanumeric codes to their medical meanings.

\section{Experimental Setup}

\subsection{Datasets}

\begin{figure}[t]
    \centering
    \includegraphics[width=\columnwidth]{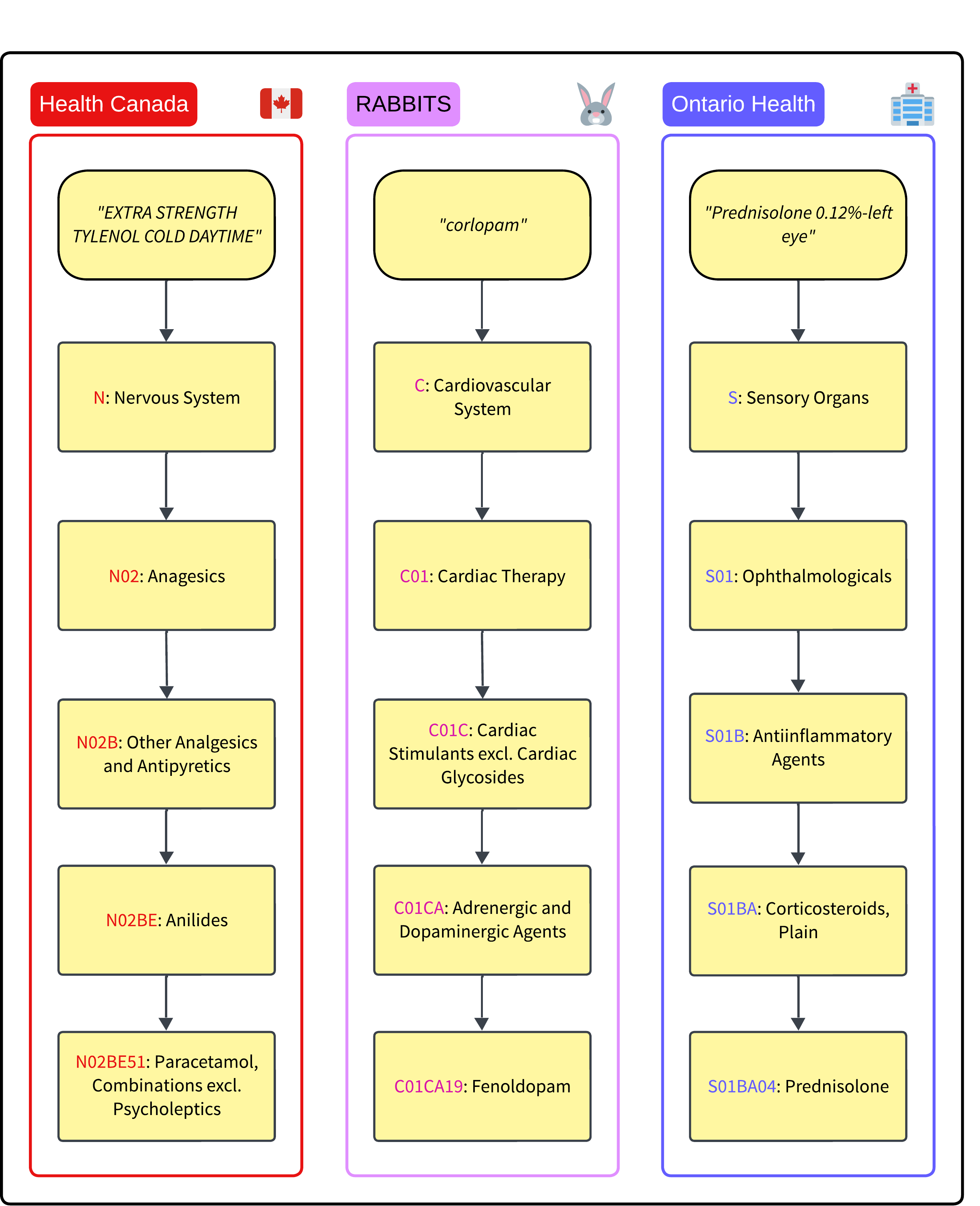}
    \caption{Examples of drug mentions and their corresponding ATC codes at each level on the Health Canada product names, RABBITS product names, and the Ontario Health assessments. Each ATC code is followed by its generic name, as in the ``With Name'' setting.}
    \label{fig:examples}
\end{figure}

\paragraph{Health Canada Product Names.}
In clinical prescriptions, healthcare providers typically specify drugs by their brand names to facilitate patient purchasing.
To develop solutions for real-world drug management and inventory control, we utilize the Drug Product Database Data Extract from Health Canada.\footnote{\href{https://www.canada.ca/en/health-canada/services/drugs-health-products/drug-products/drug-product-database/read-file-drug-product-database-data-extract.html}{https://www.canada.ca/en/health-canada/services/drugs-health-products/drug-products/drug-product-database/read-file-drug-product-database-data-extract.html}}
This comprehensive dataset contains 5,744 pairs of product names and ATC codes representing drug products approved for use in Canada.
Examples can be found in Figure~\ref{fig:examples}.
We create stratified train--test splits (90\%/10\%) based on the 14 level-1 ATC categories to ensure representative evaluation across the entire hierarchy.

\paragraph{RABBITS Product Names.} 
We further augment our evaluation with the RABBITS dataset \citep{gallifant24:rabbits}, which provides 3,680 expert-verified  pairs of product names and ATC codes sourced from RxNorm.\footnote{\href{https://www.nlm.nih.gov/research/umls/rxnorm/index.html}{https://www.nlm.nih.gov/research/umls/rxnorm/index.html}}
The dataset was specifically designed to evaluate the robustness of LLMs in handling equivalent brand and generic drug names.
Following our approach with the Health Canada dataset, we create stratified 90\%/10\% train--test splits based on level-1 ATC categories.

\paragraph{Ontario Health Assessments.} 
We obtain 200 anonymous clinical prescription notes from InterRAI Canada, sourced from Ontario Health.
All notes were verified by an expert to contain no personally identifiable information.
Each note consists of a concise, unstructured textual description of a drug, such as ``microlax miroenema''.
Being free-form clinical text, these descriptions are inherently noisy, including misspellings and mixed instructions (e.g., ``Senna, if no BM X 2 days'', ``PEG 3350- mix with 100-250ml fluid of p'').
A domain expert (JMG) manually assigned ATC codes to these prescriptions to create gold-standard labels.
All 200 prescription notes are used for evaluation.

\subsection{Evaluation Metrics}

\begin{table*}[t]
\tableformat
\centering
\begin{tabular}{|c|ccc|ccc|cc|}
\toprule
\multirow{2}{*}{Correct Level} & \multicolumn{3}{|c|}{Health Canada} & \multicolumn{3}{c|}{RABBITS} & \multicolumn{2}{c|}{Ontario Health} \\
 & Fine-tuned 8B\textsuperscript{*} & Llama 3.1 70B & GPT-4o & Fine-tuned 8B\textsuperscript{*} & Llama 3.1 70B & GPT-4o & Fine-tuned 8B\textsuperscript{*} & Llama 3.1 70B \\
\midrule
$\geq 5$ & 60.5\% & 60.3\% & 78.4\% & 26.4\% & 19.8\% & 39.4\% & 53.1\% & 49.4\% \\
$\geq 4$ & 67.7\% & 64.7\% & 79.1\% & 32.9\% & 32.1\% & 47.8\% & 68.3\% & 67.5\% \\
$\geq 3$ & 78.3\% & 74.6\% & 84.3\% & 43.5\% & 43.5\% & 55.4\% & 85.2\% & 83.2\% \\
$\geq 2$ & 84.7\% & 80.7\% & 87.3\% & 46.7\% & 52.7\% & 64.4\% & 88.3\% & 88.0\% \\
$\geq 1$ & 90.3\% & 87.1\% & 90.3\% & 62.8\% & 71.2\% & 81.8\% & 91.2\% & 89.8\% \\
\bottomrule
\end{tabular}
\caption{Accuracy at each ATC level (A@L1 through A@L5) for different LLMs, tested on Health Canada data, RABBITS product names test sets, and the 166 Ontario Health Assessments with Level 5 granularity. Each row shows the percentage of predictions at or above that correct level. Fine-tuned 8B\textsuperscript{*} refers to our fine-tuned Llama 3.1 8B. Experiments here were conducted in the ``With Name'' setting.}
\label{tab:health_rabbits_ontario}
\end{table*}

\begin{table*}[t]
\tableformat
\centering
\begin{tabular}{|c|ccc|ccc|}
\toprule
\multirow{2}{*}{Correct Level} & \multicolumn{3}{c|}{Health Canada} & \multicolumn{3}{c|}{RABBITS} \\
 & Code Only & With Name & With UMLS & Code Only & With Name & With UMLS \\
\midrule
$\geq 5$ & 40.0\% & 60.3\% & 61.2\% & 8.4\% & 19.8\% & 20.4\% \\
$\geq 4$ & 55.3\% & 64.7\% & 65.2\% & 25.0\% & 32.1\% & 32.3\% \\
$\geq 3$ & 68.7\% & 74.6\% & 74.3\% & 41.3\% & 43.5\% & 44.0\% \\
$\geq 2$ & 81.4\% & 80.7\% & 80.3\% & 53.0\% & 52.7\% & 53.3\% \\
$\geq 1$ & 89.2\% & 87.1\% & 87.1\% & 72.0\% & 71.2\% & 71.2\% \\
\bottomrule
\end{tabular}
\caption{Comparison of cumulative correct prediction levels across different knowledge grounding settings using Llama 3.1 70B. Each row shows the percentage of predictions at or above that level.}
\label{tab:grounding}
\end{table*}

\paragraph{Correct Level.}
The ATC coding system uses a hierarchical structure where each level is represented by a specific number of characters: levels 1 to 5 use 1 character, 3 characters, 4 characters, 5 characters, and 7 characters, respectively.
For level $k \in \{1, \cdots, 5\}$, let $\ell_k$ denote the number of characters used at level $k$.
Then, given an unstructured drug mention $x$, its gold label ATC code $y$, and a predicted ATC code $\hat y$, we define the \textit{correct level} of the prediction as the maximum $k \in \{1, \cdots, 5\}$ where $y$ and $\hat y$ have a common prefix of length $\ell_k$.

\paragraph{Granularity Level.}
Clinical prescriptions in the Ontario Health dataset often contain inherent ambiguities that make it challenging to confidently assign exact level-5 ATC codes, even for domain experts.
To account for this uncertainty, we introduce a \textit{granularity level} annotation ranging from 0 to 5 for each prescription text.
This metric represents the deepest level in the ATC hierarchy that can be confidently determined without ambiguity, annotated by domain expert (JMG).
For example, the prescription text ``digestive enzyme - 1 tablet'' can be classified as A09AA enzyme preparations (level-4), but lacks sufficient detail to determine the specific chemical substance (level-5), and therefore has a granularity level of 4.

When evaluating predictions for a prescription text with granularity level $k$, we consider the correct level to be at most $k$, as predictions beyond this level cannot be reliably assessed.

This granularity annotation is unique to the Ontario Health dataset, reflecting the real-world ambiguity in clinical prescriptions.
In contrast, the Health Canada product names are all assigned complete level-5 ATC codes, and the RABBITS dataset has been curated by \citeauthor{gallifant24:rabbits} to only contain unambiguous product names.

\subsection{LLM Backbones}

We evaluate two prominent LLMs in zero-shot settings:\ GPT-4o representing proprietary models,\footnote{\href{https://openai.com/index/gpt-4o-system-card/}{https://openai.com/index/gpt-4o-system-card/}} and Llama 3.1 70B representing open-source models.\footnote{\href{https://huggingface.co/meta-llama/Llama-3.1-70B-Instruct}{https://huggingface.co/meta-llama/Llama-3.1-70B-Instruct}}
Additionally, to explore more resource-efficient solutions, we fine-tune Llama 3.1 8B on the combined training sets from Health Canada product names and RABBITS.\footnote{\href{https://huggingface.co/meta-llama/Llama-3.1-8B-Instruct}{https://huggingface.co/meta-llama/Llama-3.1-8B-Instruct}}
We fine-tune at a learning rate of 2e-5 over 3 epochs with batch size 4.
All experiments maintain consistent parameters with temperature 0.1 and random seed 42.

\subsection{Knowledge Grounding Settings}
We conduct ablation experiments across three knowledge grounding settings to evaluate their impact on coding accuracy, varying the context provided for each option in the level-by-level prompt presented in Section~\ref{methods:level_prompt}:
\begin{itemize}
    \item \textbf{Code Only}:\ LLMs select from options presenting only the alphanumeric ATC codes (e.g., ``A12AA01'')
    \item \textbf{With Name}:\ Options include both the alphanumeric ATC code and its  generic name (e.g., ``A12AA01: calcium phosphate'')
    \item \textbf{With UMLS}: Options include the alphanumeric ATC code augmented with its UMLS definition, as detailed in Section~\ref{grounding}
\end{itemize}
These ablation experiments are conducted with the zero-shot models.
The Llama 3.1 8B is fine-tuned and evaluated only in the ``With Name'' setting to maintain consistent training and testing conditions.

\section{Results and Discussion}

\paragraph{Zero-shot Effectiveness.}
Table~\ref{tab:health_rabbits_ontario} presents our models' effectiveness across the three datasets, in the ``With Name'' setting.
On the Health Canada product names, GPT-4o demonstrates strong zero-shot effectiveness, achieving 78.4\% accuracy at level 5 (exact ATC code matches), while open-source Llama 3.1 70B achieves 60.3\% zero-shot.
This effectiveness gap narrows at level 4---a granularity level still commonly used in clinical research.

However, on the RABBITS dataset, while the relative effectiveness between models remains, the overall accuracy decreases by approximately 40\% compared to Health Canada results.
This effectiveness gap can be attributed to string similarity differences between product names and their corresponding generic names.
In the Health Canada dataset, 43.0\% of product names are either substrings of their generic names or vice versa, compared to only 1.4\% in RABBITS.
This disparity reveals that the zero-shot ability in LLMs to code product names stems from pre-trained knowledge of generic drug names rather than understanding of product names themselves.
When product names share less lexical similarity with their generic counterparts, the models' effectiveness degrades significantly.

For the Ontario Health assessments, among the 200 clinical prescription notes, 166 (83\%) are assigned granularity level 5, indicating that a precise level-5 ATC code can be deduced with confidence.
The remaining 34 notes are distributed across other granularity levels: 20 at level 0, 0 at level 1, 2 at level 2, 2 at level 3, and 10 at level 4.
We evaluate the open-source Llama 3.1 models (excluding GPT-4o due to privacy constraints) on the 166 unambiguous samples.
The results are on par with the Health Canada product names, particularly at correct level $\geq 4$.
This validates our hypothesis that real-world drug prescriptions are often variations of product names, and suggests that GPT-4o would likely achieve similar zero-shot effectiveness on the Ontario Health assessments as observed with the Health Canada product names.

\paragraph{Fine-tuning Effectiveness}
Notably, when fine-tuned on the Health Canada and RABBITS training sets, Llama 3.1 8B consistently surpasses the zero-shot accuracy of the larger Llama 3.1 70B model across all three datasets.
This demonstrates that effective ATC coding is possible with smaller, locally deployable models when task-specific training data is available.

\paragraph{Knowledge Grounding Effectiveness.}
Table~\ref{tab:grounding} illustrates the effect of different knowledge grounding settings using Llama 3.1 70B on the two product names datasets. 
We observe two phenomena: (1) Though below the ``With Name'' setting, the ``Code Only'' setting achieves meaningful accuracy, indicating pre-existing knowledge of ATC codes in LLMs. (2) UMLS definition grounding provides modest improvements over generic name grounding, particularly at level 5, suggesting that the additional contextual information enable the LLM to make finer decisions deeper in the ATC hierarchy, where the possible ATC codes are very similar.

\section{Conclusion}

In this work, we present a practical approach to automatic ATC coding using LLMs, demonstrating meaningful zero-shot effectiveness on both curated product-name datasets and real clinical prescriptions.
Further, we show that fine-tuned smaller models can achieve comparable effectiveness, showcasing the potential of automated ATC coding with limited computational resources.

Our analysis reveals several important insights for real-world deployment.
First, the similarity in effectiveness between Ontario Health prescriptions and Health Canada product names suggests that drug mentions in prescription settings often appear as variations of product names, where our approach demonstrates strong zero-shot accuracy.
Second, our investigation of knowledge grounding demonstrates that while additional context can improve fine-grained classification at deeper levels, the improvements are modest overall.
Finally, the effectiveness gap between Health Canada and RABBITS datasets highlights a key limitation: current LLMs rely heavily on string similarity between product names and their generic counterparts, suggesting an area for future improvement.

Looking ahead, in addition to addressing the string similarity challenge, several directions could enhance the practical utility of our system.
Developing more efficient knowledge grounding strategies could improve accuracy without sacrificing speed, and exploring hybrid approaches that combine LLM-based classification with traditional rule-based systems might provide more robust solutions for healthcare organizations.

To conclude, our work demonstrates the feasibility of automated ATC coding with LLMs, while also setting the groundwork for building careful systems that balance accuracy, privacy, and computational requirements in healthcare settings.

\section*{Acknowledgments}

This research was supported in part by the Natural Sciences and Engineering Research Council (NSERC) and InterRAI Canada.
Additional funding is provided by Microsoft via the Accelerating Foundation Models Research program.

\bibliography{zero_shot_atc}

\end{document}